\documentclass[runningheads]{llncs}
\usepackage{caption} 
\usepackage{tabularx}
\usepackage{float}
\usepackage{array}  
\usepackage{booktabs} 
\usepackage{multirow}
\usepackage{makecell}
\usepackage{arydshln} 
\usepackage{todonotes}
\usepackage{eurosym}

\usepackage{graphicx}

\begin{document}

\title{Tactile interaction with a robot leads to increased risk-taking}

\author{Qiaoqiao Ren \and Tony Belpaeme}

\authorrunning{Q. Ren et al.}

\institute{IDLab -- imec, Ghent University, Ghent, Belgium\\
\email{\{Qiaoqiao.Ren, Tony.Belpaeme\}@UGent.be}\\
}

\maketitle              
\begin{abstract}

Tactile interaction plays a crucial role in interactions between people. Touch can, for example, help people calm down and lower physiological stress responses. Consequently, it is believed that tactile and haptic interaction matter also in human-robot interaction. We study if the intensity of the tactile interaction has an impact on people, and do so by studying whether different intensities of tactile interaction modulate physiological measures and task performance. We use a paradigm in which a small humanoid robot is used to encourage risk-taking behaviour, relying on peer encouragement to take more risks which might lead to a higher pay-off, but potentially also to higher losses. For this, the Balloon Analogue Risk Task (BART) is used as a proxy for the propensity to take risks. We study four conditions, one control condition in which the task is completed without a robot, and three experimental conditions in which a robot is present that encourages risk-taking behaviour with different degrees of tactile interaction. 
The results show that 
both low-intensity and high-intensity tactile interaction increase people's risk-taking behaviour. However, low-intensity tactile interaction increases comfort and lowers stress, whereas high-intensity touch does not. 


\end{abstract}
%
%
%

\section{Introduction}

In general, research on haptic and tactile interaction with robots has received little attention in Human-Robot Interaction (HRI). Haptic interaction is prominent in Human-Computer Interaction (HCI) as an interaction modality with digital media \cite{lee2006physically}, but physical interaction with robots is rare for several reasons \cite{campeau2016time}. One is that most robots are unsafe to touch or handle, even robots which have been explicitly designed for HRI are often unsuitable for tactile interaction \cite{pervez2008safe}. The Nao robot, for example, the most widely used robot in HRI studies, has several pinch points which could trap fingers. A second reason is that most social robots are not explicitly designed for physical interaction. Often, physically manipulating or holding the robot at best triggers a fail-safe or at worst leads to breakages. Finally, we have only recently started understanding more about the possible effects of tactile interaction with robots, and this paper aims to make a contribution to understanding the effect of intensity of tactile interaction on people's attitudes and task performance with a robot.



The influence of physical interaction, and specifically touch, has been widely documented. For example, light physical contact between people has been shown to promote a sense of security and consequently increases risk-taking \cite{hri1}. Recent research found that slow, emotive touch can help reduce social suffering associated with exclusion. Slower, emotive touch --as opposed to rapid, neutral touch-- influences the perception of physical pain and appears to be mediated by a distinct tactile neurophysiological system \cite{hri4}. Also in HRI, touch and physical interaction is known to change the behaviour or perception of the user. For example, touching a PARO robot can reduce pain perception and reduce salivary oxytocin levels \cite{hri5}.  

The study reported here focuses on risk-taking behaviour. It has been established that taking risks is a fundamental human activity with significant ramifications for one's finances, health, and social life. Peer pressure, notably, can play a significant role in risk-taking behaviours. If someone sees friends or peers engaging in risky behaviour and seeks acceptance, they are more likely to engage in those behaviours. People also readily respond to encouragement, for example, drivers will drive fast when encouraged by others, which leads to a higher possibility of an accident \cite{hri9,hri10}. As such, peer influence and peer pressure play an important role in risky behaviour.

In this study, we focus on whether touching or holding a robot can change human risk-taking behaviour. Earlier work found that a robot's verbal encouragement could help people take more risks and pointed out potential implications for human decision-making \cite{hri2}. This ties in with earlier work on peer pressure, which showed that people under certain circumstances conform to robots' pressure, as a way to gain social approval \cite{vollmer2018children} or reduce informational uncertainty \cite{salomons2018humans}. However. It is yet unclear whether \emph{human-robot touch} can influence the peer pressure from robots. 



\section{Methodology}
 
\subsection{Experimental design}

\begin{figure}[!t]
\setlength{\abovecaptionskip}{-0.0cm}
\setlength{\belowcaptionskip}{-0.5cm}
\centering
\includegraphics[width=0.8\textwidth]{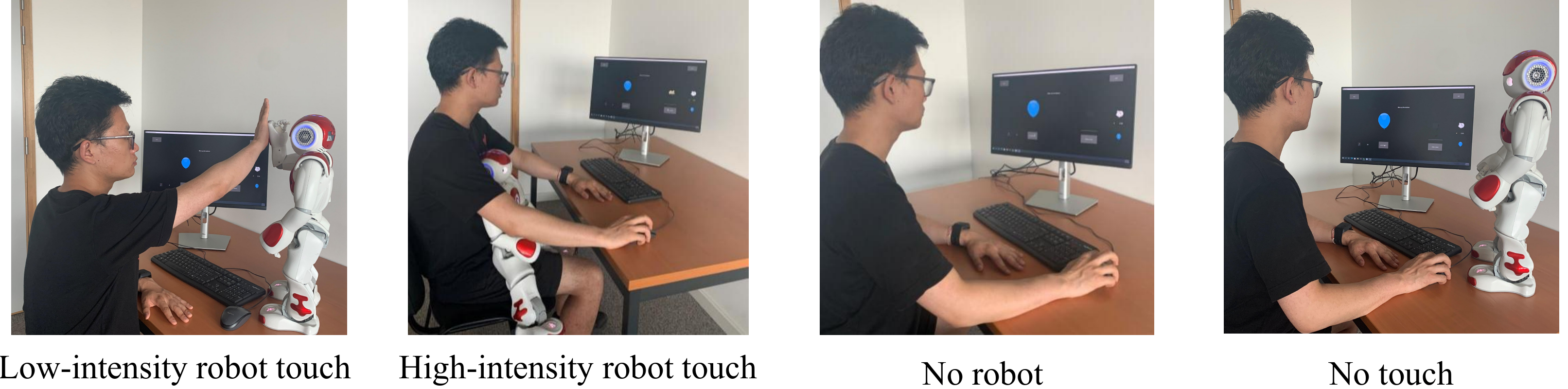}
\caption{The four interaction conditions used in the study.} \label{fig5}
\end{figure}

Participants interacted with a Softbank Robotics NAO robot. In order to assess the participant's risk-taking behaviour, a Balloon Analogue Risk Taking (BART) was used, identical to \cite{hri2}. In this, participants are shown a balloon on screen and are asked to inflate it by pressing the space bar. Their monetary earnings increase with the size of the balloon and they can collect their money by moving on to the next balloon. If the balloon explodes --which happens at a pseudorandom number of pumps-- their earnings are lost for that balloon. Each participant inflates 30 balloons. 
Four experimental conditions were used, as illustrated in Fig.~\ref{fig5}.

\begin{enumerate}
    \item In the \emph{low-intensity condition (LI)} the participant has brief and low-force tactile interactions with the robot, timed to coincide with the participant wanting to move on to the next balloon or when the robot encourages them to take more risk. At these times, the robot invites the participant to give a high-five, shake hands or touch the robot's head. 
    \item In the \emph{high-intensity condition (HI)} the participant has prolonged and high-force tactile interaction with the robot, as the robot asks the participant to be taken on their lap. As such, both the participant and the robot face the screen together. During the experiment, the robot uses its arms to point and gesticulate, just as in the previous condition.
    \item In the \emph{no-touch condition (NT)} the robot only encourages participants to take more risks by verbally encouraging them to pump more. Note that the verbal interaction between all three experimental conditions is identical. 
    \item In the \emph{no-robot condition (NR)} no robot is present and the participants play the BART games without encouragement to take more risks. This serves as a control condition and baseline to compare the experimental data against.

\end{enumerate}

We used a within-subject design, with all participants being exposed to all four conditions. The order in which we present the conditions is balanced across participants.

\begin{figure}[!b]
\setlength{\abovecaptionskip}{-0.0cm}
\setlength{\belowcaptionskip}{-0.1cm}
\centering
\includegraphics[width=0.8\textwidth]{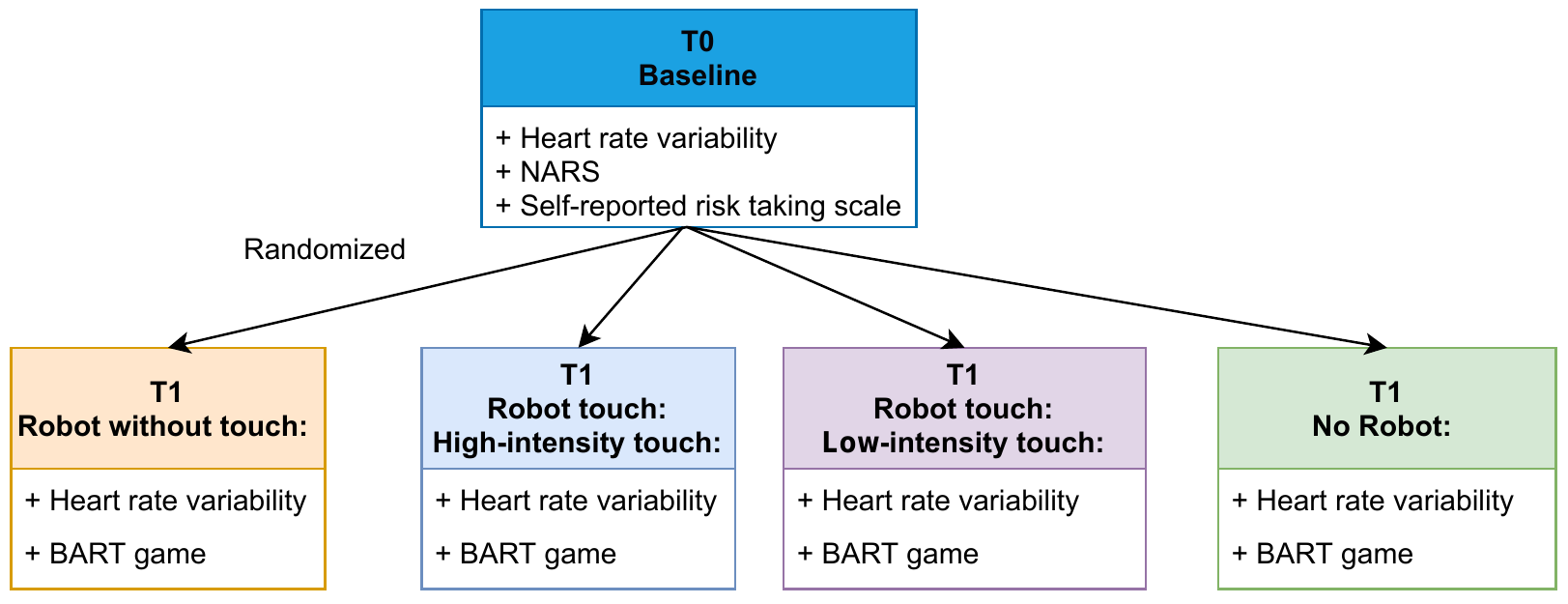}
\caption{Experiment set-up at T0.} \label{fig1}
\end{figure}




\subsection{Participants}

38 participants (19 female, 19 male; 27.0 ± 2.2 years old) were randomly distributed over four conditions: the low-intensity touch, the high-intensity touch, the no-touch and the no-robot conditions. Participants were recruited through a local social media campaign and were offered a 5 \euro voucher for an online store. Participants were excluded if they self-reported learning difficulties related to reading or hearing, or if they reported an acute or chronic heart condition, as this might impair our ability to collect appropriate data. The data collection and study adhered to the ethics procedures of the \emph{Universiteit Gent} and participants gave informed consent.

The required sample size was calculated using G*Power \cite{hri3}. Assuming an effect size of 0.23 and $\alpha = 0.05$, the statistical power is 81.5\% and 28 participants are needed. We recruited 38 participants in the experiment, which for a sample size of 38 individuals and $\alpha = 0.05$ means that the statistical power is 87.2\%.

\subsection{Measurements}

\subsubsection{BART score}




The \emph{BART score} is used as a measure of the participant's risk-taking behaviour. It is equal to the  the adjusted average number of pumps on unexploded balloons \cite{bornovalova2005differences,lejuez2002evaluation}, with higher scores indicative of greater risk-taking propensity. Other measures which can be informative are the number of exploded balloons, and participants’ profits for each balloon. The BART score, the number of pumps, exploded balloons, and profits are summed over the 30 trials \cite{hri2}.



\subsubsection{Heart rate variability}
We also collect physiological data, as a proxy for stress, measuring the inter-beat interval (IBI) using the Empatica E4 sensor. 
The E4 sensor comes as a wristband and was worn on the non-dominant hand. Participants were asked to refrain from moving that hand, so as not to generate spurious readings. 

The heart rate variability (HRV) reveals variations between consecutive inter-beat intervals and is a good measure of emotional arousal. Other measures indicative of physiological components associated with emotional regulation, such as the Parasympathetic Nervous System (PNS) index, the Sympathetic Nervous System (SNS) index, and Stress (the square root of Baevsky’s stress index \cite{baevsky2008methodical}), can be calculated from the inter-beat interval data using the Kubios HRV software. 

Heart rate variability is known to increase in response to parasympathetic nervous system (PNS) activity (vagal stimulation) and decrease with 
sympathetic nervous system (SNS) activity \cite{hri7,hri8,hri6}. HRV is influenced by stress and current neurobiological evidence supports HRV as an objective measure of psychological health and stress. The high-frequency component of HRV is reduced when a subject is under stress or experiences emotional strain \cite{kim2018stress}. Earlier research indicated that HRV increases when subjects perceive prolonged tactile stimulation as pleasant \cite{triscoli2017heart}. Likewise in this study, we explore how the intensity of touch affect stress levels.

\subsubsection{Self-reported risk-taking propensity}

Subjective self-reported risk-taking behaviour was measured by asking participants to rate a single question (``How do you see yourself? Are you generally a person who is fully prepared to take risks or do you try to avoid taking risk'') on a 7-point scale, from 1 (not at all willing to take risks) to 7 (very willing to take risks). 


\subsubsection{Negative attitudes towards robots}

The Negative Attitude Toward Robots Scale (NARS) is a measure designed to gauge how individuals feel and behave when interacting with various types of robots. Three subscales make up the NARS: negative attitudes toward situations of interaction with robots, negative attitudes toward the social influence of robots, negative attitudes toward emotions in interaction with robots. There are 17 items total, and each is evaluated from 1 (strongly disagree) to 5 (strongly agree).\cite{nomura2006measurement} \cite{tsui2010using}


\subsection{Procedure}


Four BART games, each with 30 trials, were played by each participant. Each participant experienced all four conditions --low-intensity touch, high-intensity touch, no-touch, no-robot-- with the order of the conditions being randomly balanced across participants. The participants were instructed to wear the Empatica E4 sensor on their non-dominant hand, as identified by the experimenter during the completion of the informed consent form. Before the first BART task, participants completed a NARS questionnaire and self-reported risk-taking scale and were made to relax for at least 15 minutes to get a baseline for the IBI data recorded by E4 sensor. After each BART game, another IBI measurement is taken (at times T1, T2, T3, T4). The procedure illustrated in Fig.~\ref{fig1} shows one possible unfolding after T1. Upon finishing, participants were debriefed and asked for feedback. Participants took approximately 1 hour and 10 minutes to complete the study. 



\subsection{Data analysis}


Data processing and analyses were done using Python and IBM SPSS V28 (IBM, Armonk, NY, USA), IBI data were analysed using Kubios HRV standard and then analyzed by SPSS. A one-way repeated measures ANOVA was used to compare data between conditions. A Shapiro-Wilk analysis (with $\alpha = 0.05$) was used to confirm the normality of distributions. Outliers were identified were removed if they were more than 5 standard deviations removed from the mean. Moreover, all data meet the Sphericity assumption according to Mauchly’s Test of Sphericity (with $\alpha = 0.05$). The  Mauchly’s W and $p$ value are reported in Table~\ref{tab5}. 



\section{Experimental results and analysis}\label{AA}


\subsection{BART performance}


The performance on the BART task is reported using the \emph{BART score}. Due to the low sample size ($N < 40$), a Shapiro-Wilk test was used to test for normality of the distribution of the results. 

Based on this outcome, and after visual inspection of the histogram (frequencies for intervals of values of a metric variable) and the Normal Quantile-Quantile plot (which compares the observed experiment data quantiles with the  quantiles that we would expect to see if the data were normally distributed) for BART scores in four conditions, a parametric test was used in this experiment. The results of the Shapiro-Wilk test and Mauchly's test are reported in Table~\ref{tab4} and Table~\ref{tab5} respectively.

A one-way repeated analysis of variance (ANOVA) indicates significant differences in BART scores between the four interaction conditions, $F(3, 38) = 4.70, p < 0.05$. Post hoc Tukey's HSD tests (with Bonferroni corrections) revealed that there are significant differences between the low-intensity touch condition and no-touch condition ($p=0.019;p < 0.05$) and between the high-intensity touch and no robot touch conditions ($p=0.007;p < 0.05$). See Fig.~\ref{fig3} and Table~\ref{tab1}.

\begin{table}[t]\footnotesize 
\setlength{\abovecaptionskip}{0.0cm}   
	\setlength{\belowcaptionskip}{-0cm}  
	\renewcommand\tabcolsep{2.0pt} 
	\centering
	\caption{Test of Normality for BART scores and heart rate variability.}
	\begin{tabular}
	{
	p{2.5cm}<{\centering} |
 p{2cm}<{\centering} |
	 p{2.3cm}<{\centering} |
  p{2.2cm}<{\centering}|
	p{2.2cm}<{\centering}
	} 
\hline
    
     {Shapiro-Wilk} & 
     {HI scores}  & 
     {LI scores} & 
     {NT scores} & 
     {NR scores}  \\
     
     \hline

   {BART Scores}&
  {0.98, $p > 0.05$} &
 {0.95, $p > 0.05$} &
 {0.98, $p > 0.05$} &
 {0.97, $p > 0.05$} \\

     \hline

   {PNS index}&
  {0.95, $p > 0.05$} &
 {0.99, $p > 0.05$} &
 {0.97, $p > 0.05$} &
 {0.96, $p > 0.05$}\\
     \hline
     
     {SNS index}&
  {0.97, $p > 0.05$} &
 {0.97, $p > 0.05$} &
 {0.97, $p > 0.05$} &
 {0.98, $p > 0.05$}\\
     \hline
     
     {Stress index}&
  {0.89, $p > 0.05$} &
 {0.95, $p > 0.05$} &
 {0.90, $p > 0.05$} &
 {0.89, $p > 0.05$}\\
     \hline
     
	\end{tabular}
	\label{tab4}
\end{table}

\vspace{-.2cm}

\begin{table}[h]\footnotesize 
\setlength{\abovecaptionskip}{0.0cm}   
	\setlength{\belowcaptionskip}{-0cm}  
	\renewcommand\tabcolsep{2.0pt} 
	\centering
	\caption{Mauchly's Test of Sphericity for BART scores and heart rate variability.}
	\begin{tabular}
	{
	p{2.5cm}<{\centering} |
 p{2cm}<{\centering} |
	 p{2.3cm}<{\centering} |
  p{2.2cm}<{\centering}|
	p{2.2cm}<{\centering}
	} 
\hline
    
     \multirow{2}{*} {Mauchly's W} & 
     {BART Scores}  & 
     {PNS index} & 
     {SNS index} & 
     {Stress index}  \\

     \cline{2-5}

   &
  {0.78, $p > 0.05$} &
 {0.92, $p > 0.05$} &
 {0.83, $p > 0.05$} &
 {0.97, $p > 0.05$} \\

     \hline
     
	\end{tabular}
	\label{tab5}
\end{table}

\vspace{-1.0cm}

\begin{figure}[h!]
\setlength{\abovecaptionskip}{0.2cm}
\setlength{\belowcaptionskip}{-0.6cm}
\centering
\includegraphics[width=0.6\textwidth]{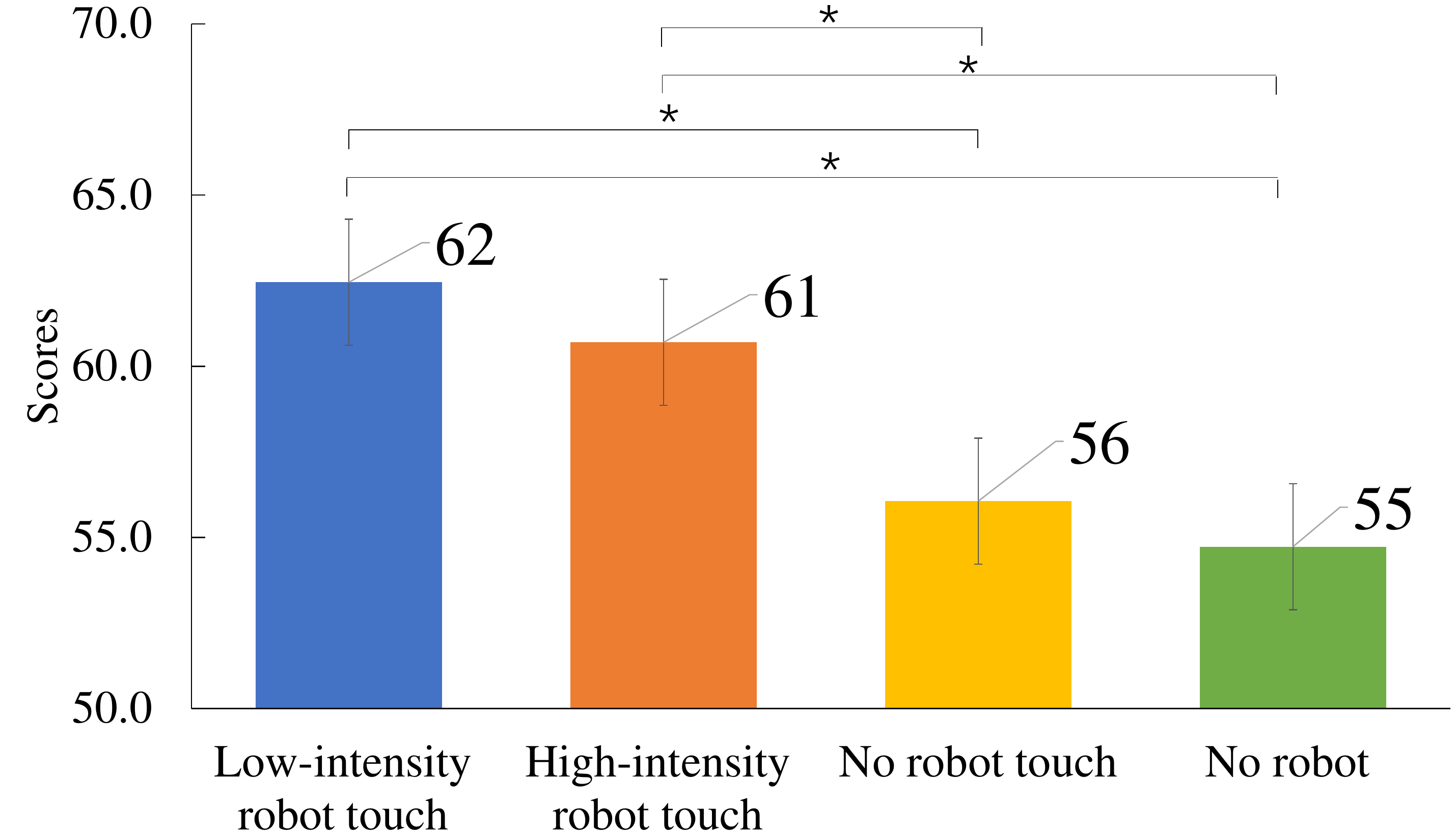}
\caption{BART scores per condition. Bars show the average scores for each condition, whiskers indicate standard error. (* $p < 0.05$).} \label{fig3}
\end{figure}

\vspace{-0.6cm}

\subsection{Heart rate variability}

\begin{figure}
\setlength{\abovecaptionskip}{-0.2cm}
\setlength{\belowcaptionskip}{-0cm}
\includegraphics[width=\textwidth]{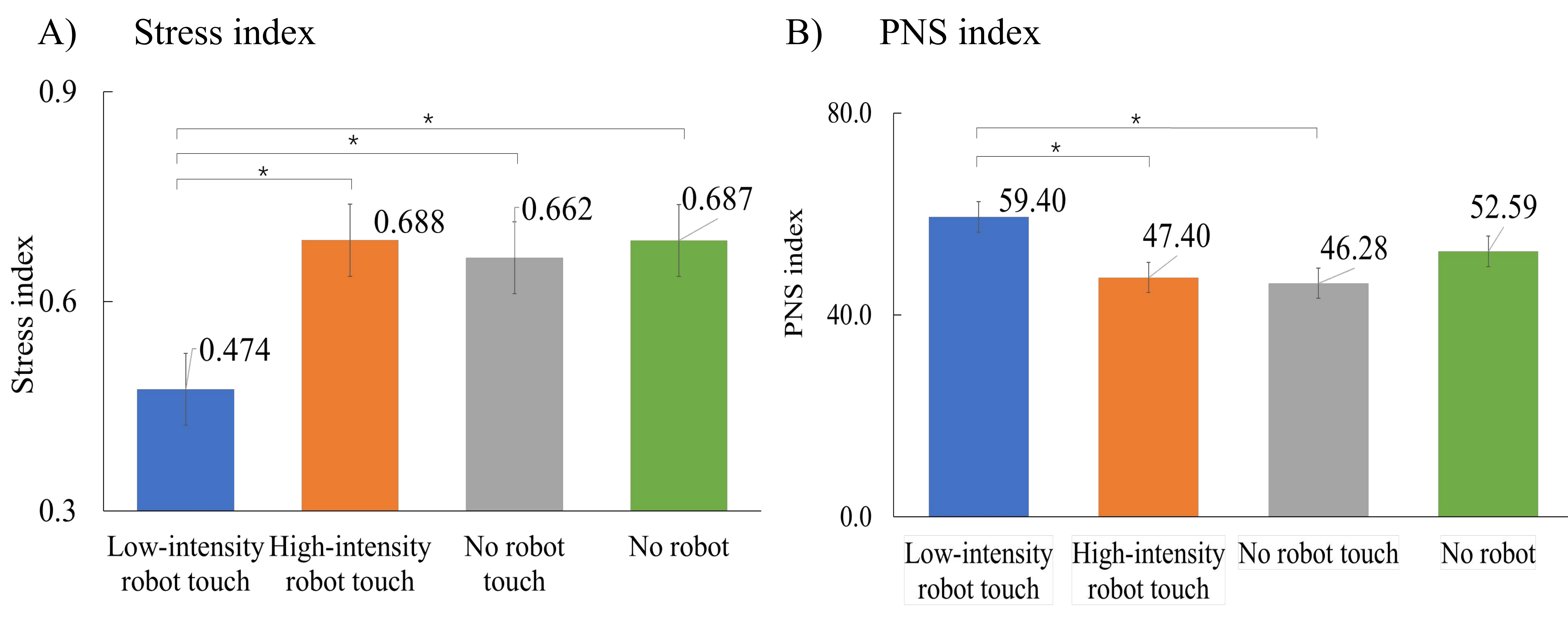}
\caption{The mean Stress (left) and PNS (right) indices across the four conditions. Lower values for Stress and higher values for PNS indicate the participant experiences less stress. Whiskers indicate standard error. ($*p < 0.05, **p < 0.01$).} \label{fig4}
\end{figure}

As mentioned earlier, heart rate variability is a good measure to analyse stress levels and emotional arousal. However, HRV is often indirectly reported through other measures, such as the PNS, SNS and Stress index. 
When stressed, participants are expected to have lower PNS and higher SNS index values. From Fig.~\ref{fig4} and Table~\ref{tab1} we can see that in the low-intensity touch condition, the highest PNS index is obtained, which means the participants have the least stress in this condition. Participants have higher stress in high-intensity touch and no-touch interactions, according to the Stress index. The repeated ANOVA shows that the PNS index significantly differs between conditions ($F(3,36)=4.531, p < 0.05$). Post hoc tests on the PNS index (with Bonferroni corrections) revealed that there are significant differences between the low-intensity and no-touch conditions, and between the low-intensity touch and high-intensity touch. 
The Stress index also shows a significant effect of condition ($F(3,36)=5.65, p < 0.05$). Post-hoc tests (with Bonferroni corrections) revealed that there are significant differences between no touch, low-intensity touch and high-intensity touch. The above suggests that participants felt more pressure in HI and NT conditions whereas less pressure was experienced by them in LI condition.

\vspace{-1.2cm}

\begin{table}\footnotesize 
\setlength{\abovecaptionskip}{0.0cm}   
	\setlength{\belowcaptionskip}{-0cm}  
	\renewcommand\tabcolsep{2.0pt} 
	\centering
	\caption{Means (and Standard Deviations) of Self-Reported Risk Taking, negative attitudes towards robots (NARS) and BART scores by Condition.}
	\begin{tabular}
	{
	p{2cm}<{\centering} |
 p{1.2cm}<{\centering}
	 p{1.cm}<{\centering} |
  p{1.2cm}<{\centering}
	p{1.cm}<{\centering} |
 p{1.2cm}<{\centering}
	p{1cm}<{\centering} |
	p{1.2cm}<{\centering} 
 p{1cm}<{\centering} 
	} 
\hline
        \multirow{2}{*}{Variable} &
        \multicolumn{2}{c|}{High-intensity} &
        \multicolumn{2}{c|}{Low-intensity} &
        \multicolumn{2}{c|}{No-touch} &
        \multicolumn{2}{c}{\multirow{2}{*}{No-robot}} \\

        & \multicolumn{2}{c|}{touch} & \multicolumn{2}{c|}{touch} & \multicolumn{2}{c|}{touch} & \\

          \hline
          
	Metric	& M & SD
  & M & SD  
  & M & SD 
  & M & SD \\

    \cline{1-9}
     {BART Scores}  & 61 & 12 &
     62 & 12 &
     56 & 13 &
     55 & 14 \\
     \hline

     {Stress index}  & 0.69  & 0.35  &
    0.47  & 0.3  &
     0.66 & 0.34 &
     0.69  & 0.32 \\
     \hline

     {PNS index}  & 47.40 & 3.24 &
    59.40 & 3.37 &
    46.28 & 3.49 &
    52.59  & 4.71 \\
     \hline

     {SNS index}  & -2.83 & 0.20 &
     -3.09 & 0.11 &
     -2.79 & 0.18 &
      -3.01& 0.19 \\
     \hline
	\end{tabular}
	\label{tab1}
\end{table}

\vspace{-1.0cm}

\subsection{Correlation analysis}

The correlation between self-reported risk-taking and BART scores is calculated using Pearson’s $r$.
There were significant positive correlations in all conditions ($p < 0.01$), which indicates that the participants' performance over four conditions is consistent. If they have higher BART scores in one condition, then higher BART scores are shown in other conditions. The correlation between total BART scores over the whole experiment and the self-reported risk taking scores is positive and significant ($r = 0.437, p < 0.01$), meaning that more risk-seeking people have higher BART scores. 



The correlation between Negative Attitude Toward Robots Scale (NARS) and BART scores is shown in Table~\ref{tab3}. There is a significant negative correlation between no-touch BART scores, and NARS scores ($p < 0.05$), which suggests that participants who have a negative attitude towards the robot take less risk. There is no significant correlation between no-robot BART scores and NARS scores, which makes sense as no robot was present to influence risk-taking behaviour. Remarkably this correlation is greater in conditions where participants have a tactile interaction with the robot as there is a greater negative correlation between BART scores and NARS scores ($p < 0.01$).










\begin{table}[h!]\footnotesize 
\setlength{\abovecaptionskip}{0.3cm}   
	\setlength{\belowcaptionskip}{-0.3cm}  
	\renewcommand\tabcolsep{2.0pt} 
	\centering
	\caption{Correlations between NARS scores and BART scores including the HI scores, LI scores and NT scores, NR scores, which means the BART score during HI, LI, NT and NR conditions respectively.	* means the correlation is significant at $p < 0.05$, ** at $p < 0.01$.}

	\begin{tabular}
	{
	p{3cm}<{\centering} |
 p{1.8cm}<{\centering} |
	 p{1.8cm}<{\centering} |
	 	 p{1.8cm}<{\centering} |
  p{1.8cm}<{\centering}
	} 
\hline
    
    & 
     {HI scores}  & 
     {LI scores}  & 
     {NT scores}  & 
     {NR scores} \\
\cline{2-5}
 {Negative attitudes towards robots} & 
 {-0.49**, p\textless0.01} &
  {-0.64**, p\textless0.01} &
 {-0.34*, p\textless0.05} &
 {-0.19,  $p > 0.05$} \\

     \hline
	\end{tabular}
	\label{tab3}
\end{table}


\vspace{-0.8cm}

\section{Discussion}

Earlier research indicated that a verbal robot encourages participants to do more risk-taking behavior. \cite{hri2}, a result which we could not replicate in our study. However, our study extends earlier work by showing that tactile interaction with a robot increases risk-taking behaviour, but also reduces the stress felt during risk-taking: both the low-intensity and high-intensity tactile interaction promote risk-taking behavior. In contrast, non-tactile interaction with the robot has no significant influence on participants' risk-taking behavior.

When comparing the low-intensity and high-intensity tactile interaction, there are significant differences in experienced stress (as indicated by the PNS, SNS, and Stress index). Participants experience less stress in the low-intensity tactile interaction, and most stress when the robot is present but not being touched or when no robot is present. These results conclusively show that tactile interaction has a beneficial impact on stress and suggest that low-intensity tactile interaction might be preferred over high-intensity tactile interaction. Low-intensity touch seems to make people respond emotionally and reduces stress. A majority of participants (90\%) mentioned that holding the robot on their lap in the HI condition felt somewhat like holding a baby and the robot --at 5.5 kg-- felt rather heavy after a while. They felt a responsibility towards the robot, which might show in the stress measures. In the LI condition, participants indicated that they experienced the robot as encouraging and comforting.

As expected, self-report risk-taking scores are correlated with BART scores. 
Moreover, the Negative Attitude Towards Robot score (NARS) negatively correlated with the BART scores in the no-touch interaction, which means that the participants feel less peer pressure from the robot if they hold a more negative attitude towards robots. Moreover, during tactile interaction, the negative correlation is stronger than when the robot is not being touched or held, suggesting that tactile interaction softens any negative preconceptions towards robots.


Another, perhaps relevant, factor is that in the low-intensity condition the robot is standing on a table facing the participant and looking slightly down at the participant. In the high-intensity condition the robot is sitting on the lap and facing away from the participant. Eye gaze and size do matter in HRI \cite{kompatsiari2017importance} \cite{admoni2017social}, so this might have an impact on the outcomes of social human-robot interaction, something which future studies would need to confirm.


In this study, we examined that human-robot tactile interaction could increase human risk-taking behaviour, and low-intensity tactile interaction could help humans decrease stress during risky occasions. On the one hand, our research alarms that robots might be causing potential threats by increasing risk-taking behaviour. On the other hand, there is the possibility of utilizing the human-robot tactile interaction  in psychological treatment in hospitals for patients or preventing loneliness at home for those who live alone.

\vspace{-0.1cm}

\bibliographystyle{unsrt}

\bibliography{references.bib}




\end{document}